\documentclass[a4paper, 10 pt, epsfig, conference]{ieeeconf}

\IEEEoverridecommandlockouts                              % This command is only needed if 
\usepackage{float}
\usepackage[dvipdfmx]{graphicx}
\usepackage{latexsym}
\usepackage{url}
\usepackage{cite}
\usepackage{comment}
\usepackage{algorithm}
\usepackage{algpseudocode}

\begin{comment}
\date{
\begin{list}{}{
\setlength{\leftmargin}{20mm}
\setlength{\rightmargin}{20mm}
}
\setlength{\baselineskip}{2.0mm} %4.5mm
\item {\normalsize
\vspace{-13mm}
\hspace{-1.5mm}
\vspace{-1mm}\\
\hspace{5mm}
}
\end{list}
\vspace{-7mm}
}
\begin{document}
\maketitle
\setlength{\baselineskip}{4.0mm}%%行間
\thispagestyle{empty}
\renewcommand{\figurename}{Fig.}
\renewcommand{\tablename}{Table}
\end{comment}

\setlength{\topmargin}{9mm}

\begin{document}

\newcommand{\FIG}[3]{
\begin{minipage}[b]{#1cm}
\begin{center}
\includegraphics[width=#1cm]{#2}\\
{\scriptsize #3}
\end{center}
\end{minipage}
}

\newcommand{\FIGU}[3]{
\begin{minipage}[b]{#1cm}
\begin{center}
\includegraphics[width=#1cm,angle=180]{#2}\\
{\scriptsize #3}
\end{center}
\end{minipage}
}

\newcommand{\FIGm}[3]{
\begin{minipage}[b]{#1cm}
\begin{center}
\includegraphics[width=#1cm]{#2}\\
{\scriptsize #3}
\end{center}
\end{minipage}
}

\newcommand{\FIGR}[3]{
\begin{minipage}[b]{#1cm}
\begin{center}
\includegraphics[bb=0 0 640 480,angle=-90,width=#1cm]{#2}
\\
{\scriptsize #3}
\vspace*{1mm}
\end{center}
\end{minipage}
}

\newcommand{\FIGRpng}[5]{
\begin{minipage}[b]{#1cm}
\begin{center}
\includegraphics[bb=0 0 #4 #5, angle=-90,clip,width=#1cm]{#2}\vspace*{1mm}
\\
{\scriptsize #3}
\vspace*{1mm}
\end{center}
\end{minipage}
}

\newcommand{\FIGpng}[5]{
\begin{minipage}[b]{#1cm}
\begin{center}
\includegraphics[bb=0 0 #4 #5, clip, width=#1cm]{#2}\vspace*{-1mm}\\
{\scriptsize #3}
\vspace*{1mm}
\end{center}
\end{minipage}
}

\newcommand{\FIGtpng}[5]{
\begin{minipage}[t]{#1cm}
\begin{center}
\includegraphics[bb=0 0 #4 #5, clip,width=#1cm]{#2}\vspace*{1mm}
\\
{\scriptsize #3}
\vspace*{1mm}
\end{center}
\end{minipage}
}

\newcommand{\FIGRt}[3]{
\begin{minipage}[t]{#1cm}
\begin{center}
\includegraphics[angle=-90,clip,width=#1cm]{#2}\vspace*{1mm}
\\
{\scriptsize #3}
\vspace*{1mm}
\end{center}
\end{minipage}
}

\newcommand{\FIGRm}[3]{
\begin{minipage}[b]{#1cm}
\begin{center}
\includegraphics[angle=-90,clip,width=#1cm]{#2}\vspace*{0mm}
\\
{\scriptsize #3}
\vspace*{1mm}
\end{center}
\end{minipage}
}

\newcommand{\FIGC}[5]{
\begin{minipage}[b]{#1cm}
\begin{center}
\includegraphics[width=#2cm,height=#3cm]{#4}~$\Longrightarrow$\vspace*{0mm}
\\
{\scriptsize #5}
\vspace*{8mm}
\end{center}
\end{minipage}
}

\newcommand{\FIGf}[3]{
\begin{minipage}[b]{#1cm}
\begin{center}
\fbox{\includegraphics[width=#1cm]{#2}}\vspace*{0.5mm}\\
{\scriptsize #3}
\end{center}
\end{minipage}
}

\title{\Large \bf
Active Robot Vision for Distant Object Change Detection: \\
A Lightweight Training Simulator Inspired by Multi-Armed Bandits
}

\author{Terashima Kouki$^{1}$, Tanaka Kanji$^{1}$, Yamamoto Ryogo$^{1}$, and Tay Yu Liang Jonathan$^{1}$
%\thanks{*This work was not supported by any organization}% <-this % stops a space
\thanks{$^{1}$K. Terashima, K. Tanaka, R. Yamamoto, and J. Tay Yu Liang are with Robotics course, Faculty of Engineering, University of Fukui, Japan. {\tt\small \{ha209059, tnkknj, mf220362, mf228029\}@g.u-fukui.ac.jp}}%
}

\date{}

\maketitle

\newcommand{\figA}{
\begin{figure}
\begin{center}
\vspace*{5mm}
\FIG{8.5}{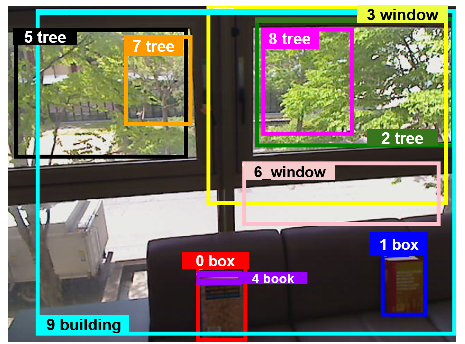}{}
\caption{Distant objects are often misrecognized as unknown objects (i.e., appearance change) due to their low resolution.
% Objects at distance are often misrecognized as change objects.
}\label{fig:A}
\end{center}
\end{figure}
}

\newcommand{\figB}{
\begin{figure}
\begin{center}
\vspace*{5mm}
\FIG{8}{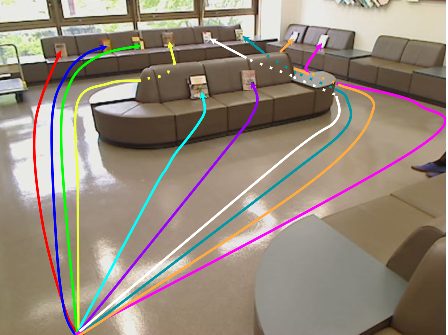}{}\\
\FIG{8}{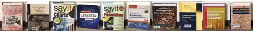}{}
\caption{Viewpoint trajectories from a low-level action planner (top) and map object images (bottom).}\label{fig:B}
\end{center}
\end{figure}
}

\newcommand{\figC}{
\begin{figure}
\vspace*{5mm}
\FIG{8}{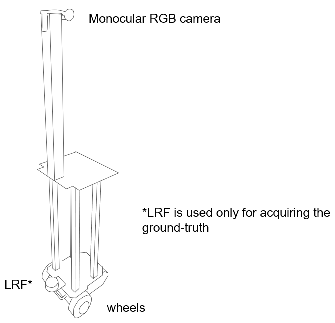}{}\vspace*{-3mm}\\
\caption{An active robot vision system.}\label{fig:C}
\end{figure}
}

\newcommand{\figD}{
\begin{figure}
\begin{center}
\vspace*{5mm}
\FIG{8}{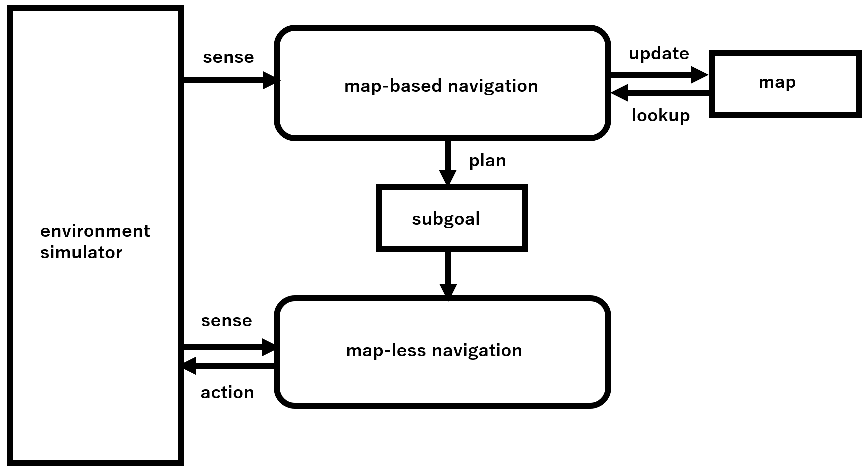}{}~\vspace*{5mm}\\
\caption{Hierarchical planner.}\label{fig:D}
\end{center}
\end{figure}
}

\newcommand{\figE}{
\begin{figure}
\begin{flushleft}
\FIG{8}{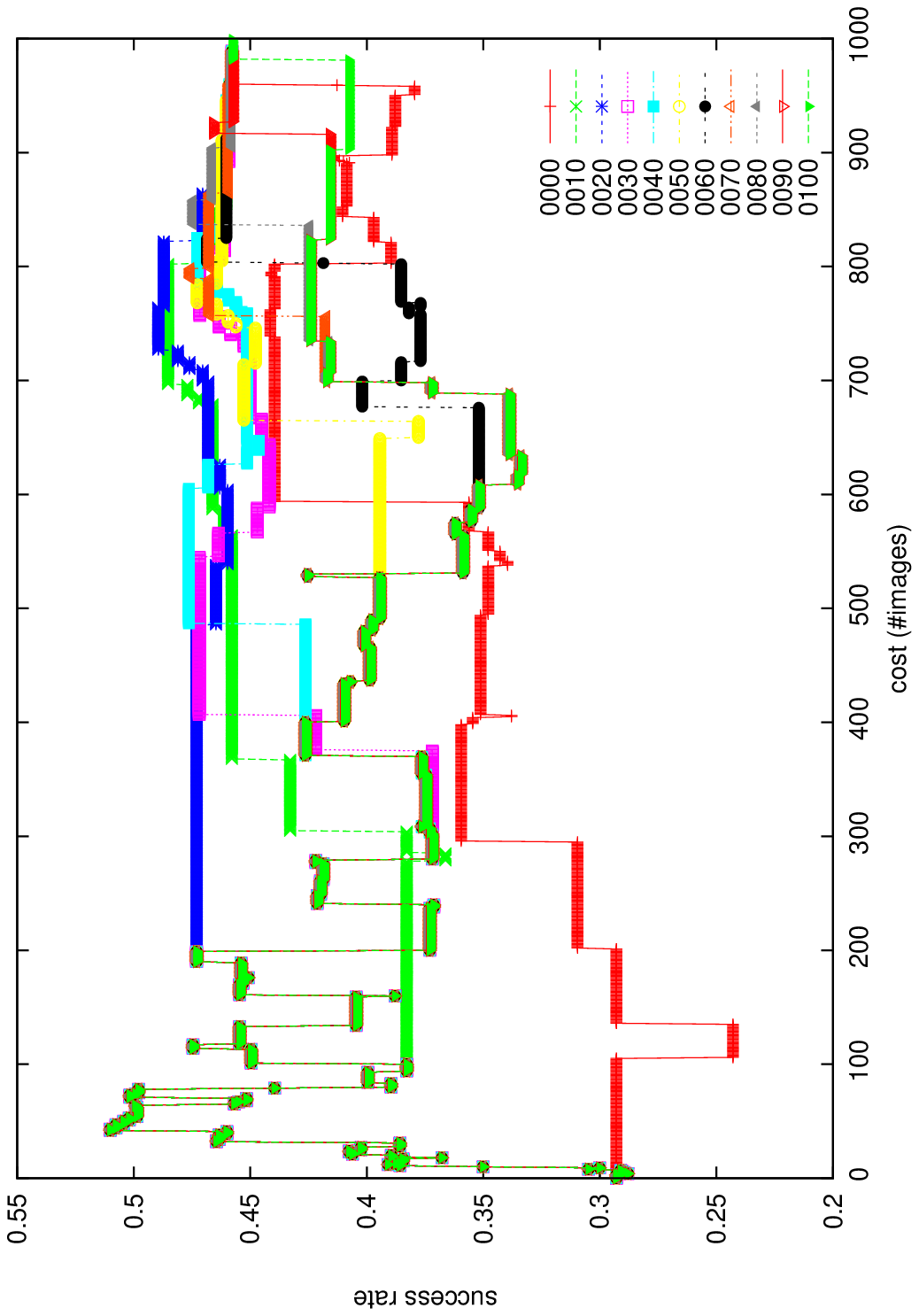}{}
\caption{Performance results. Vertical axis: detection accuracy. Horizontal axis: navigation cost. Legend: hyperparameter $L'$.}\label{fig:E}
~\vspace*{-7mm}\\
\end{flushleft}
\end{figure}
}

\newcommand{\figF}{
\begin{figure}
\begin{center}
~\vspace*{3mm}\\
\FIG{8}{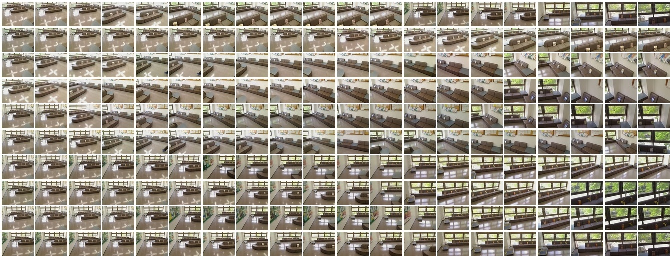}{}\vspace*{-3mm}\\
% ~\vspace*{3mm}\\
% \FIG{16}{fig6.eps}{}\vspace*{-3mm}\\
\caption{Different journeys often look similar to each other, which may confuse the action planner. 
The panel sequence in each row is samples of images from each journey's raw image sequence, arranged from left to right in ascending order of timestamps.
}\label{fig:F}
\vspace*{-5mm}
\end{center}
\end{figure}
}

\newcommand{\figG}{
\begin{figure}
\begin{flushleft}
\FIG{8}{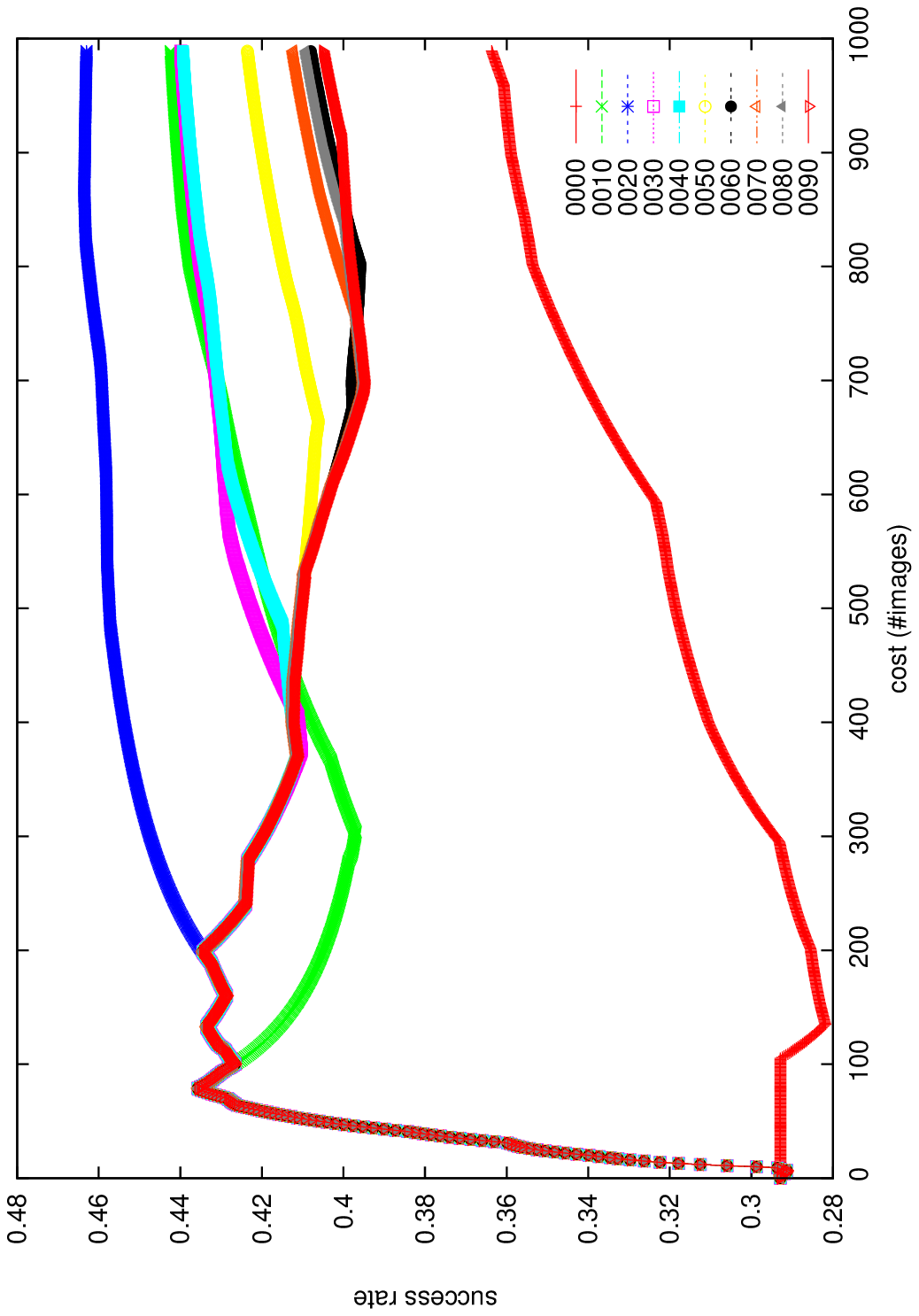}{}
\caption{Average performance versus available budget.}\label{fig:G}
\end{flushleft}
\end{figure}
}

\begin{abstract}
In ground-view object change detection, the recently emerging mapless navigation has great potential to navigate a robot to objects distantly detected (e.g., books, cups, clothes) and acquire high-resolution object images, to identify their change states (no-change/appear/disappear). However, naively performing full journeys for every distant object requires huge sense/plan/action costs, proportional to the number of objects and the robot-to-object distance. To address this issue, we explore a new map-based active vision problem in this work: ``Which journey should the robot select next?" However, the feasibility of the active vision framework remains unclear; Since distant objects are only uncertainly recognized, it is unclear whether they can provide sufficient cues for action planning. This work presents an efficient simulator for feasibility testing, to accelerate the early-stage R\&D cycles (e.g., prototyping, training, testing, and evaluation). The proposed simulator is designed to identify the degree of difficulty that a robot vision system (sensors/recognizers/planners/actuators) would face when applied to a given environment (workspace/objects). Notably, it requires only one real-world journey experience per distant object to function, making it suitable for an efficient R\&D cycle. Another contribution of this work is to present a new lightweight planner inspired by the traditional multi-armed bandit problem. Specifically, we build a lightweight map-based planner on top of the mapless planner, which constitutes a hierarchical action planner. We verified the effectiveness of the proposed framework using a semantically non-trivial scenario ``sofa as bookshelf".
\end{abstract}

\section{
Introduction
}

Ground-view change detection plays an important role for a robot to detect inconsistencies between the environment model (``map") and live scenes, in emerging application domains such as lifelong learning \cite{l3} and long-term autonomy \cite{lta}. Unlike typical vision setups such as parallel projection satellite imagery \cite{satellite}, the problem is complicated by non-linear perspective projection and visual uncertainties (e.g., depth ambiguity, aspect differences, occlusion) \cite{gvcd}.

\figA

One of the open issues in ground-view change detection is distantly detected object changes. Since distant objects (e.g., books, cups, clothes) are typically observed as spatially low-resolution object images, they often look significantly different than they were seen during training and are misrecognized as change even with state-of-the-art object detection models. An example of such misrecognition is shown in Fig. \ref{fig:A}, where a book is misrecognized as a box. 

In our observation, the recently emerging point-goal navigation \cite{pgn} and other mapless navigation techniques \cite{huang2023visual,ogn,vln,ans} have great potential to navigate a robot to distant objects and acquire high-resolution object images, to identify their change states (no-change/appear/disappear). Specifically, these recent navigation techniques act as a reasonably efficient and lightweight action planner for purely vision-based long-distance navigation, which has traditionally been considered a computationally difficult problem.

Nevertheless, the feasibility of such an active vision framework remains uncertain: (1) Those planners do not address online domain adaptation, as they rely on offline deep reinforcement learning which requires a huge dataset and time costs; and (2) These planners are not capable of selecting an appropriate goal object but only navigating to a given goal. In particular, the lack of object selection capability is a serious issue. This is because naively visiting every distant object requires a large amount of sense/plan/action cost that is proportional to the number of objects and the robot-to-object distance. Unfortunately, it is a non-trivial task to select an appropriate goal object among distant objects that are typically recognized uncertainly \cite{dod}. Note that it is not even clear whether we can do better than a naive action planner that randomly samples one out of the distant objects \cite{yamauchi1997frontier}.

To address this issue, we present an efficient simulator for feasibility testing, intending to accelerate the early-stage R\&D cycles (e.g., prototyping, training, testing, evaluation). The proposed simulator is designed to identify the degree of difficulty that a robot vision system (sensors/recognizers/planners/actuators) would face when applied to a given environment (workspace/objects). This simulator requires a very small amount of real-world data, only one journey per distant object, to function, making it suitable for an efficient R\&D cycle. It exploits the analogy between our selection problem ``Which object-specific journey should the robot select next?" and the traditional multi-armed bandit problem ``Which arm should the player select next?".

Another contribution of this work is that we introduce a lightweight map-based planner on top of mapless navigation, which constitutes a hierarchical planner (Fig.\ref{fig:D}) that combines the advantages of the map-based planner (i.e., domain-adaptability, goal selection capability) and the mapless planner (i.e., domain-invariance, goal navigation). The effectiveness of the proposed framework has been experimentally verified using the semantically non-trivial scenario ``sofa as bookshelf".

\figD

\subsection{Discussions}

The feasibility test addressed by our simulator aims to identify the degree of difficulty that an active vision system (sensors/recognizers/planners/actuators) would face when deployed in a given environment (workspace/objects). For example, a test result showing that an active vision system has the same level of performance as a naive random action planner is an indicator of the fact that the application is too difficult for that system. As a practical example, in the experimental section of this paper (Section \ref{sec:exp}), we present the results of a feasibility test on the real-world system we are developing, where the system significantly outperforms a random action planner.

Our simulator requires very little real-world data, only one mapless journey per distant object, to function. This is in contrast to the typical sim-to-real approach (addressed also in our parallel work \cite{acpr2023yoshida}) that requires rich real-world data from the target workspace for reconstructing the simulated environment. Furthermore, this simulator does not require detailed geometric models of the workspace, which are not available in mapless navigation.

\section{
Related Work
}

The problem of image change detection has been studied in various formulations. One popular formulation is a setup of pairwise image comparison for remote sensing applications \cite{satellite}, in which detecting changes across different domains (e.g., across seasons) based on comparisons between query and map images is a topic of ongoing research \cite{gvcd}. However, these studies rely on simplified vision setups such as parallel projection views. This is not the case for ground-view change detection, where the nonlinear perspective projection further complicates the problem. This complex and ill-posed scenario has recently been investigated with support from high-capacity deep learning models \cite{2d2d}. This line of research can be further categorized into 2D-3D comparison \cite{2d3d}, 3D-3D comparison \cite{3d3d}, and 2D-2D comparison \cite{2d2d}, depending on whether the input scene or map model used for comparison is 2D (e.g., monocular RGB image) or 3D (e.g., LIDAR point cloud). In particular, the 2D-2D comparison is one of the most challenging formulations, where the 3D model is not available in both the training and test stages. Our problem formulation falls into this category.

Our approach is most closely related to the research field of ground-view object change detection, aiming to maintain a ``bag-of-objects" map, which is an unordered collection of objects detected at the time of mapping within the same workspace. In \cite{he2022diff}, the problem of detecting significant changes such as the appearance and disappearance of portable traffic lights on high-definition HD maps was addressed by a two-step procedure: (1) Rasterization by projecting map elements from the camera's perspective; (2) Extraction of pyramid features with different resolutions from both the rasterized and camera images. This approach is effective for change detection in large-scale, high-resolution HD maps. Compared with these existing studies, our approach has two key novelties. First, rather than assuming passive vision as in most previous works, we consider active vision (e.g., \cite{tanaka2003viewpoint}), including the problem of next-best-view action planning, to determine distant object changes. Second, we do not rely on high-definition 3D models but use only a 2D ``bag-of-objects" image model.

Mapless navigation has several variants such as point-goal navigation \cite{pgn}, object-goal navigation \cite{ogn}, and vision-language navigation \cite{vln}. Among them, point goal navigation is most relevant to our problem. Formally, it is a workspace-specific lightweight state-to-action map that aims to map a visual input to a next-best-view action. However, it relies on costly offline deep learning and thus is not available for fast online adaptation. In other words, it is assumed that the traversability of the workspace does not change significantly. Such an assumption is relevant in our indoor mobile robot application, where object changes often occupy only a small portion of the input image, and they do not significantly affect the mapless navigation capability. In this work, we aim to explore the mapless navigation framework from a new perspective of map-based active vision.

\section{Dataset}\label{sec:dataset}

Figure \ref{fig:C} shows an overview of the robot. A front-facing onboard camera is used for change detection and collision avoidance. At the time of writing, the mapless navigation module of our robot is under development, and therefore, in the current experiments, it was run in a human-in-the-loop manner with experimenter assistance. Additionally, an LRF was used as an auxiliary proximity sensor for collision avoidance and for recording the ground-truth viewpoint trajectories. 

Figure \ref{fig:B} shows an overview of the ``sofa as bookshelf" scenario. Although our framework can be generalized to various types of mapped objects, in this work, we consider a specific experimental scenario in which a book is the only object type, to simplify the experiments and analysis. Thus, books with different colors, textures, and sizes are recorded in the ``bag-of-objects" format \cite{sbow}. Now, the purpose of active change detection here is to identify whether each book is unchanged. Note that in cases where a book turns out to be changed, we need to consider a post-process of classifying the book's change state into ``appearance" (e.g., the book was occluded by the newly added objects) or ``disappearance" (e.g., the book was removed from the workspace). However, such a change state classification task is out of the scope of the current work and left for our future research. 

We observed the active change detection task to be very challenging. First, none of the objects were similar to the target object image, nor were they even correctly recognized as books. Since sofas were habitually used as book storage in this workspace, books were often misrecognized as pillows, which are semantically more compatible than books with sofas. Second, sofas were often not categorized as sofas. One reason might be that chairs, a superordinate category of sofas, are a difficult object class to recognize even with the support of affordance cues \cite{grabner2011makes}.

\section{Approach}\label{sec:approach}

We formulate the active vision problem as reinforcement learning \cite{suttonrl}, a standard framework for training an action planner. In contrast to other frameworks like supervised and unsupervised learning, the model (i.e., agent) is trained via interaction with the environment in reinforcement learning. Simulated environments are commonly used in robot navigation tasks to reduce the risk of damaging the robot and the cost of robot-environment interaction. For example, in our parallel work \cite{acpr2023yoshida}, we used a photo-realistic physics simulator Habitat. Such simulators are very useful for training environment-independent action planners. However, as reported in recent publications, a model trained in a simulated workspace may have poor performance when deployed in a very different real-world workspace. One straightforward way to address this issue is to build a 3D photo-realistic simulated environment that is equivalent to the target workspace. However, this requires a rich training set acquired in the target workspace and a high-grade 3D reconstruction technique, which significantly limits its applications.

\figC

\figB

Our simulator is introduced to overcome the limitations of these existing simulators. This proposed simulator has several desirable properties. First, it requires only a simple map consisting of $N$ objects that are distantly detected at a certain viewpoint. Second, it does not require rich datasets but only requires one real-world journey per distant object to function. Third, it does not assume the availability of 3D reconstruction technology. This simulator is specifically described as follows.
The internal state of the simulator at timestep $t$ is represented by an $N$-dim state vector $s^t=$$(s^t[1], \cdots, s^t[N])$, where $s^t[n]$ indicates the unseen part $V_n(s^t[n]), \cdots, V_n(L)$ of each $n$-th journey $V_n$ at timestep $t$. 
At simulator startup $t=0$, the state vector is initialized to $s^0=(1, \cdots, 1)$. At each timestep $t$, the planner is required to select an appropriate journey $n^*$ ($n^*\in [1, N]$) followed and subsequently sends a ``one step forward" action request $n^*$ to the simulator.
Then, the simulator returns the first image $V_{n^*}(s^t[n^*])$ in the unseen part of the $n^*$-th journey, followed by the removal of that image ID from the unseen part (i.e., $s^{t+1}[n^*]$$\leftarrow$$s^t[n^*]+1$). 
Note that since each object-specific journey is initialized to length $L$, the robot is not allowed to make more than $L$ action requests per journey. If the top-priority action request by the robot is such an unacceptable action (i.e., $s^t[n^*]=L$), the next-priority action request will be executed instead.
In experiments, the number of distant objects was $N=10$, and each journey consists of a length $L=100$ image sequence, which constitutes 
(1) a size $NL$ image set. Figure \ref{fig:F} shows sample images from the journeys.
As shown, different journeys often look similar to each other, which may confuse an active vision system.

Our active change detection framework
exploits the analogy between the journey selection problem ``Which journey should the robot choose next?" and the multi-armed bandit problem ``Which arm should the player choose next?."
Specifically, the active vision framework is a double-loop algorithm. 
The outer loop of this algorithm is a map-based navigation phase,
while the inner loop is a mapless navigation phase.
We focus on the map-based navigation phase, assuming a standard implementation of mapless navigation.
One execution of map-based navigation consists of two steps: 
(1) Evaluate the similarity between each distant object and the target object;
(2) Determine ``Which distant object should the robot navigate to?" based on the similarity score. 
% Note that a mapless navigation model could predict the cost of reaching an object through simulation. However, this prediction is often inaccurate. In the current experiments, this cost was simply approximated by the number of images acquired. An example implementation of active and passive vision will be shown in the subsections below.
In this way, the map-based planner module aims to determine the order by which the distant objects should the robot navigate to. 

There are two things to note. 
First, the exact navigation cost (i.e., sensing, recognition, planning, and action costs) required by an object-specific journey cannot be known in advance, but it can only be estimated by the black-box mapless planner module. In the current experiments,  we simply approximate this cost by only the number of image acquisitions (i.e., only the sensing cost). 
Second, meaningful measurements can only be obtained when the target object is observed closely. In other words, no meaningful measurement can be obtained in cases 
where other objects are observed
or 
where the target object is observed at a distance. 
Therefore, 
a straightforward approach to address this issue would be to navigate the robot toward the most likely object with the highest similarity score
at each map-based planning phase.

\figF

\subsection{Active Vision}

Our active vision approach exploits the analogy between our problem and the ``multi-armed bandit (MAB)" problem.
In MAB the player is required to select one out of $N$ arms, whereas in our problem a robot is required to select one out of $N$ journeys at each timestep.
In MAB the focus of the exploration phase is to learn the expected reward for each arm in MAB, whereas in our problem it is to learn the expectation that the target object will be detected in each journey. 
In MAB the focus of the exploitation phase is to choose an arm that is expected to maximize the reward, whereas in our problem it is to choose a journey where the expected chance of detecting the target object is maximized.

A simple MAB algorithm that runs the exploration and exploitation phases in sequence is considered. If an agent (i.e., robot) can exploit the current knowledge of the environment, it might be able to identify which part of the environment is worth exploring. However, acquisition of such knowledge requires exploring the environment beforehand. 

Therefore, a key hyperparameter $L'$ is introduced to control the balance between the budgets for exploitation and exploration. 
Given the hyperparameter $L'$, the exploration phase is simply reduced to the task of performing the $L'$ actions in a breadth-first manner for every journey. 
This will result in obtaining $NL'$ images.
On the other hand, the exploitation phase is the task of executing the journeys in a depth-first manner, in an order prioritized by the expected rewards. 

At the beginning of the exploitation phase, the top-priority journey with the highest expected reward is selected. 

In the exploitation phase, the robot iterates to select the top-priority journey $n^*$ with the highest expected reward 
\begin{equation}
n^*=\arg \max_{\{n: s^t[n]<L\}} S(n)
\end{equation}
among non-completed journeys followed by execution of the selected journey with length $s^t[n^*]$, until the budget is exhausted. Here, $S(\cdot)$ is a given similarity function, one example of which will be illustrated in Section \ref{sec:implementation}.

\subsection{Passive Vision}

The passive vision module has two roles. One is to keep up-to-date the environmental knowledge: ``Which of the distantly detected objects is most likely to be the target object?". This is done by evaluating the similarity of each journey's object to the target object within an image embedding space. Another role is to transfer cues to the action planner that addresses the journey selection problem: ``Which of the distantly detected objects should the robot navigate to?". Detailed implementation issues will be discussed in the next section.

\section{Implementation}\label{sec:implementation}

Our approach is implemented in a reinforcement learning framework. 
We leave the detailed theory and terminology of reinforcement learning to literature (e.g., \cite{suttonrl}), and focus on clarifying the component correspondence between reinforcement learning and our approach.

The action space is a collection of distant object IDs $n\in [1, N]$ each of which corresponds to the action plan ``The robot navigates to the $n$-th distant object'' or ``The robot selects $n$-th journey". 

The state vector 
was defined as 
the number of image acquisitions for each journey:
$s^t=$$(s^t[1], \cdots, s^t[N])$.

The state transition is expressed as
\begin{equation}
s^{t+1}[n] \leftarrow 
\left\{
\begin{array}{ll}
s^t[n]+1 & (n=n^*) \\
s^t[n] & (\mbox{Otherwise})
\end{array}
\right.
,
\end{equation}
when the robot moves forward on the $n^*$-th journey at timestep $t$ and subsequently acquires an image $V_{n^*}(s^t[n^*])$ at the new viewpoint.
The cost is simply approximated by the sensing cost, 1 per acquired image, independent of the executed action $n^*$, as mentioned earlier.

% A simplified active vision task consisting of two phases, exploration and exploitation, was considered. The exploration phase aims to learn the expected rewards equally for all distant objects by performing a predefined number of $L'$ journeys driven by mapless navigation on all distant objects. 
% Note that the knowledge acquired here consists of $NL'$ images. 

The expected reward of a distant object at a certain point in time is evaluated as the degree of similarity between the detected object and the target object image with the best imaging conditions among the previously observed images. In our case, this is given by the formula below.
\begin{equation}
S(n) = \max_{j\in[1, s^t[n]-1]}
\cos(C(V_q), C(V_n(j))). \label{eqn:A}
\end{equation}
The function $C(\cdot)$ is an image embedding using the convolutional neural network vgg16 \cite{vgg16}, and returns the signal vector of the 13th layer given a query image input to vgg16.

% Every time a journey is completed, the exploitation planner module is called and it selects the journey with the highest expected reward at that time as the next journey, except for journeys that have already been completed.

\section{
Experiments
}\label{sec:exp}

The performance of the proposed active vision framework has been experimentally evaluated using the dataset and methodology explained in Sections \ref{sec:dataset}, \ref{sec:approach}, and \ref{sec:implementation}.

%A key design parameter for reinforcement learning-based planners is how to establish a good balance between the budgets for exploitation and exploration. 
%
Recalling that in our MAB setup the balance can be controlled by the hyperparameter $L'$. Increasing $L'$ means giving more weight to exploration, whereas reducing $L'$ means putting more weight on exploitation. We repeated experiments for different settings of the hyperparameter $L'$ to evaluate the influence of $L'$ on the performance.  

Top-1 accuracy is used as the main performance index. It is defined as the ratio of successful test samples. A test sample consists of a size $N=10$ set of length $L=100$ image sequences randomly sampled from the $N=10$ real-world image sequences, and (2) a ground-truth object randomly sampled from the $N$ map objects. For a given test sample, the active change detection task is regarded to be successful at a certain time $t \in [1, NL]$ if the distant object or journey $n^*$ $(n^*\in [1, N])$ with the highest score at that time is the same as the ground-truth object. Note that in our dataset, each real-world sequence was longer than $L=100$ in length, so each random sampling would generate a different sequence.

\figE

Figure \ref{fig:E} shows the experimental results. As shown, if the budget paid for the exploration stage $NL'$ is too small (e.g., $L'=1$), the prior knowledge of the environment in the exploitation stage is often insufficient and the performance degrades to the level of a naive random action planner. In contrast, if the budget for exploration is too large (e.g., $L'=50$), the performance of the exploitation stage could be maximized, but at the expense of increasing cost.

We conducted ablation research considering the diversity of available budgets. In general, the available budget depends on the application/user. For example, for a user robot whose primary mission is mail delivery, change detection is often a secondary task that cannot be paid much budget. In contrast, for a user robot whose main mission is map maintenance, a large amount of budget may be available for change detection. Considering such diversity in available budgets, we performed performance evaluations for different settings of available budgets. Specifically, we simulated 999 different cases, ranging from $L'=1$ to $L'=999$, and examined the user's performance for the available budget. The results are summarized in Fig. \ref{fig:G}. As shown, users characterized by $L'=20$ benefited the most from the active vision algorithm considered here. It is also noteworthy that the proposed planner proved to be significantly superior to a naive random action planner in the experiments considered here.

In conclusion, the feasibility tests considered here demonstrate that our active vision system has a sufficient level of recognition ability to select an appropriate target object among distant objects uncertainly recognized. Nevertheless, it remains unclear whether this result can be generalized to other situations that may be encountered in the workspace. For example, the performance can vary depending on various factors, such as the robot-to-object distances, the visual object recognizability, and visual foreground-background compatibility. Nevertheless, we believe that this test still serves as an efficient way to test, for a given recognizer-planner pair, whether that recognizer can provide useful cues to the planner.

\section{Conclusions}

We explored a challenging action planning problem in a distant object change detection scenario: ``Which object-specific journey should the robot navigate next?". We build a map-based planner module for goal object selection on top of the goal-oriented mapless navigation module, which constitutes a hierarchical planner. We also introduced an efficient simulator for a feasibility test, for a given distant object recognizer,  to decide whether the recognizer can provide useful cues for an action planner. We experimentally verified the effectiveness of the proposed method using a semantically non-trivial dataset ``sofa as bookshelf."

The proposed simulator extremely simplifies the lower-level planning tasks, which significantly accelerates the sense-plan-act cycle in the high-level planning task. As an additional advantage, it requires only one real-world journey experience per distant object to function, which is a particularly property for the physical robot design stage.

There are several avenues for future research. One of the avenues is to improve the proposed multi-armed bandit framework and improve its performance. Empirical verification can be performed by conducting evaluation experiments and case studies in application scenarios where distant object detection is critical (e.g., map maintenance applications). Another interesting area for study is investigating the factors that determine why and how map-based and mapless planners collaborate \cite{ans} to improve overall performance. Finally, research on the effectiveness of specific strategies to transfer the findings in the proposed lightweight planner to more sophisticated but expensive planners (e.g., deep reinforcement learning planners \cite{ktrl}) would be an important direction of future research.

\figG

\bibliography{reference} %hoge.bibから拡張子を外した名前
\bibliographystyle{unsrt} %参考文献出力スタイル

\end{document}